\documentclass[11pt,a4paper]{article}
\usepackage[hyperref]{acl2021}
\usepackage{times}
\usepackage{latexsym}

\usepackage[T1]{fontenc}

\usepackage{graphicx}

\usepackage[nameinlink]{cleveref}

\crefformat{section}{\S#2#1#3} 
\crefformat{subsection}{\S#2#1#3}
\crefformat{subsubsection}{\S#2#1#3}

\newenvironment{itemize*}%
  {\begin{itemize}%
    \setlength{\itemsep}{0.9pt}%
    \setlength{\parskip}{0.9pt}%
    \setlength{\topsep}{0.9pt}}%
  {\end{itemize}}

\usepackage{microtype}

\aclfinalcopy 

\title{Changing the World by Changing the Data}

\author{Anna Rogers \\
Center for Social Data Science \\
University of Copenhagen\\
\texttt{arogers@sodas.ku.dk} \\}

\date{}

\begin{document}
\maketitle
\begin{abstract}
NLP community is currently investing a lot more research and resources into development of deep learning models than training data. While we have made a lot of progress, it is now clear that our models learn all kinds of spurious patterns, social biases, and annotation artifacts. Algorithmic solutions have so far had limited success. An alternative that is being actively discussed is more careful design of datasets so as to deliver specific signals. This position paper maps out the arguments for and against data curation, and argues that fundamentally the point is moot: curation already is and will be happening, and it is changing the world. The question is only how much thought we want to invest into that process. 
\end{abstract}

\section{Introduction}

The key ingredient behind the recent successes in NLP is Transformer-based language models. The paradigm of pre-training followed by fine-tuning on downstream tasks was popularized by BERT \cite{DevlinChangEtAl_2019_BERT_Pre-training_of_Deep_Bidirectional_Transformers_for_Language_Understanding}, and is actively developed \cite{RogersKovalevaEtAl_2020_Primer_in_BERTology_What_We_Know_About_How_BERT_Works}. In December 2020 the human performance baselines on SuperGLUE \cite{WangPruksachatkunEtAl_2019_SuperGLUE_Stickier_Benchmark_for_General-Purpose_Language_Understanding_Systems} were surpassed twice, making the community wonder if it is possible to formulate benchmarks not solvable in this paradigm.

However, the successes are not the full story. It is becoming increasingly clear that much of the remarkable performance is down to benchmarks that do not actually require sophisticated verbal reasoning skills due to annotation artifacts and spurious patterns correlating with the target labels \cite{GururanganSwayamdiptaEtAl_2018_Annotation_Artifacts_in_Natural_Language_Inference_Data,McCoyPavlickEtAl_2019_Right_for_Wrong_Reasons_Diagnosing_Syntactic_Heuristics_in_Natural_Language_Inference,PaulladaRajiEtAl_2020_Data_and_its_discontents_survey_of_dataset_development_and_use_in_machine_learning_research}.The social biases in  NLP models are also attracting more attention \cite{ShengChangEtAl_2019_Woman_Worked_as_Babysitter_On_Biases_in_Language_Generation,DavidsonBhattacharyaEtAl_2019_Racial_Bias_in_Hate_Speech_and_Abusive_Language_Detection_Datasets,HutchinsonPrabhakaranEtAl_2020_Social_Biases_in_NLP_Models_as_Barriers_for_Persons_with_Disabilities}. 

The ``garbage in, garbage out" principle suggests that the situation will not change without a dramatic reappraisal of how NLP data is collected, both for pre-training and task-specific resources. But that seemingly uncontroversial conclusion is at the core of the interdisciplinary tension between NLP understood as a deep learning (DL) application area, and the more qualitative approaches of computational linguistics and AI ethics. How deep that tension goes is illustrated by the recent heated  (and sometimes less than professional\footnote{\href{https://www.theverge.com/22309962/timnit-gebru-google-harassment-campaign-jeff-dean}{https://www.theverge.com/22309962/timnit-gebru-google-harassment-campaign-jeff-dean}}) debate around ``On the Dangers of Stochastic Parrots: Can Language Models Be Too Big? \raisebox{-5pt}{\includegraphics[scale=0.1]{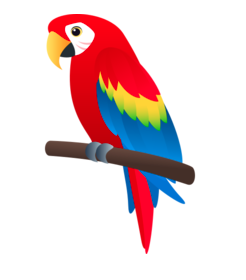}}
" by Bender, Gebru et al (\citeyear{BenderGebruEtAl_2021_On_Dangers_of_Stochastic_Parrots_Can_Language_Models_Be_Too_Big}). 

This position paper brings together the arguments for and against curating data\footnote{In this paper ``data curation" is interpreted broadly as making choices about what should be included in a NLP resource (either for pre-training or task-specific data). The phenomena to be included/excluded could be defined in terms of \textit{what} is said (e.g. soccer commentary), \textit{how} it is expressed (e.g. with or without expletives), and/or \textit{who} is speaking or being addressed (e.g. teenage soccer fans).} from linguistic and ethical perspectives (\cref{sec:debate}). It makes the case that curation is unavoidable and already happening, and that any data choices that we make, explicitly or implicitly, \textit{will} affect the real world (\cref{sec:world}). Thus the debate is only about how much thought we should put into this process. If we are to at least try to steer it, we have to overcome the interdisciplinary tension and reconsider what counts as ``NLP work'' (\cref{sec:interdisciplinary}). \Cref{sec:solutions} outlines some policies that could help.

\section{To Curate or Not to Curate?}
\label{sec:debate}

\subsection{Why Change the Data?}
\label{sec:pro}

The core argument for active curation/design of the data that goes into NLP models is that the models are representations of the data they were trained on, and thus data work is necessary to make sure that the models can learn what we need them to learn. The supporting evidence for this position comes independently from several directions: the studies finding that the models fail to learn a certain phenomenon and/or learn something undesirable.

\subsubsection{Social biases} 
\label{pro:soc}
Our world is far from perfect, and written texts contain plenty of evidence of all kinds of social biases based on gender, race, social status, ability, age, etc.  Models may learn these biases (from pre-training and/or task data) and even amplify them, putting the minority groups at a disadvantage by direct psychological harm and propagation of stereotypes \cite{BlodgettBarocasEtAl_2020_Language_Technology_is_Power_Critical_Survey_of_Bias_in_NLP,BenderGebruEtAl_2021_On_Dangers_of_Stochastic_Parrots_Can_Language_Models_Be_Too_Big}. In this context, ``data curation'' means selecting data based on its sociocultural characteristics \cite{JoGebru_2020_Lessons_from_archives_strategies_for_collecting_sociocultural_data_in_machine_learning}. Fundamentally, this is about fair representation for different social groups.

Some dismiss \citet{BenderGebruEtAl_2021_On_Dangers_of_Stochastic_Parrots_Can_Language_Models_Be_Too_Big} as ``political'', or even ``advocacy rather than research'' \cite{Lissack_2021_Slodderwetenschap_Sloppy_Science_of_Stochastic_Parrots}. However, 
``papers advocate for specific research agendas all the time'' \cite{Venkatasubramanian_2021_On_stochastic_parrots}. NLP in particular has a growing subfield of bias mitigation (see e.g. the survey on such work for gender bias by \citet{SunGautEtAl_2019_Mitigating_Gender_Bias_in_Natural_Language_Processing_Literature_Review}) that pursues exactly the same social justice agenda, but does not receive the same pushback.

\subsubsection{Privacy concerns.} Models may memorize specific facts in training data, and if those facts happen to be personally identifiable information, this is a security concern. For instance, \citet{CarliniTramerEtAl_2020_Extracting_Training_Data_from_Large_Language_Models} showed that GPT-2\footnote{Google legal department reportedly requested edits to the article by \citet{CarliniTramerEtAl_2020_Extracting_Training_Data_from_Large_Language_Models}, in particular to avoid mentions of Google technology \cite{Dave_2021_Google_pledges_changes_to_research_oversight_after_internal_revolt}.} was able to memorize personal contact information, even if it only appeared on a few web pages. A big problem is that this is not a bug, but a feature: we do want our language models to represent some facts about presidents -- just not about private citizens. Deciding what should not be remembered is clearly a data curation issue.

\subsubsection{(Lack of) progress towards NLU}

DL models are data-hungry, and so far we have heavily relied on the sources that are easy to scale: web texts for pre-training, and crowdsourcing for annotation or generating shorter texts. Combined with most funding and effort allocated to model development, this meant a less clear view of what was in the data. Consequently, the recent years witnessed a lot of findings along the following lines. 

\paragraph{DL models learn spurious patterns present in the data.} These patterns can be the results of the heuristics used by crowd workers \cite{GururanganSwayamdiptaEtAl_2018_Annotation_Artifacts_in_Natural_Language_Inference_Data}, small samples of workers creating large parts of data with traces \cite{GevaGoldbergEtAl_2019_Are_We_Modeling_Task_or_Annotator_Investigation_of_Annotator_Bias_in_Natural_Language_Understanding_Datasets}, or simply random patterns  in the task or pre-training data. For example, words like \textit{football} may frequently occur in abusive tweets, but this should not give the model the idea that all sports fans are violent \cite{WiegandRuppenhoferEtAl_2019_Detection_of_Abusive_Language_Problem_of_Biased_Datasets}. The result is that many current datasets can (and do) get ``solved" with shallow cues such as lexical co-occurrence  \cite{JiaLiang_2017_Adversarial_Examples_for_Evaluating_Reading_Comprehension_Systems,McCoyPavlickEtAl_2019_Right_for_Wrong_Reasons_Diagnosing_Syntactic_Heuristics_in_Natural_Language_Inference}. The larger the resource, the more difficult it is to avoid them  \cite{GardnerMerrillEtAl_2021_Competency_Problems_On_Finding_and_Removing_Artifacts_in_Language_Data}.

\paragraph{DL models are surprisingly vulnerable to basic perturbations.} ACL 2020 best paper award went to \citet{RibeiroWuEtAl_2020_Beyond_Accuracy_Behavioral_Testing_of_NLP_Models_with_CheckList}'s demonstration that even the successful, commercially deployed NLP systems cannot handle many core linguistic phenomena like negation. Pre-trained language models by themselves do not necessarily cope with them either \cite{Ettinger_2020_What_BERT_is_not_Lessons_from_new_suite_of_psycholinguistic_diagnostics_for_language_models}. This suggests that the current resources do not provide the signal to learn the necessary linguistic paradigms. 

\paragraph{DL models struggle to learn rare phenomena.} \label{pro:long-tail}
Linguistic phenomena generally follow Zipf distribution \cite{Zipf1945}, which means that most of them are rare in naturally occurring data, and thus harder for the models to learn. This applies even to the large pre-training datasets. For example, \citet{ZhangWarstadtEtAl_2020_When_Do_You_Need_Billions_of_Words_of_Pretraining_Data} compared the learning rates for different linguistic phenomena as RoBERTa was pre-trained on more and more data. English irregular verb forms (highly frequent) were acquired in under 10M of training tokens, but the model struggled with island effects even after 30B tokens. Such results suggest that if something needs to be learned, the model needs to be provided with a sufficiently strong signal (and it may still fail even then \cite{GeigerCasesEtAl_2019_Posing_Fair_Generalization_Tasks_for_Natural_Language_Inference}).

The bottom line is that the distributions of linguistic phenomena in the current NLP resources do not seem to provide the signal with which the current models could learn to perform human-level language ``understanding''. We do not even know the full spectrum of abilities that would qualify for that. Choosing which aspects of a given ``task'' (or language, in case of pre-training) a given resource would ``teach'' explicitly is a curation decision.

\subsubsection{Security concerns.} A relatively recent development is ``universal adversarial triggers": adversarial attacks on the models that modify the textual input in a way that forces the models to always output a certain prediction \cite{WallaceFengEtAl_2019_Universal_Adversarial_Triggers_for_Attacking_and_Analyzing_NLP}. For example, the authors make a SQuAD-trained reading comprehension model to always predict the answer ``to kill American people" for any why-question. This effect is robust and model-independent: i.e. it is the training data that gets ``hacked", not the model. 

It is not clear if it is possible to construct a dataset that would not have such vulnerabilities, but common sense suggests that the training data should be curated so as to make them unlikely to occur in the natural distribution of user input. 

\subsubsection{Evaluation methodology.} So far the fundamental paradigm for NLP work based on machine learning focused on in-distribution evaluation: the test sample would come from the same distribution as the train/validation samples, and the samples would be randomly split. Within that paradigm, it is essential that there are no overlaps between training and test data, which is an issue for many current resources \cite{LewisStenetorpEtAl_2021_Question_and_Answer_Test-Train_Overlap_in_Open-Domain_Question_Answering_Datasets,EmamiSulemanEtAl_2020_Analysis_of_Dataset_Overlap_on_Winograd-Style_Tasks}. 

To do that well, we already have to make decisions about what counts as ``overlap", and what should be in the training and testing data. For example, in pre-training GPT-3 \cite{BrownMannEtAl_2020_Language_Models_are_Few-Shot_Learners} decisions had to be made about which benchmarks would be used for evaluation. There was a (partially successful) attempt to simply remove documents with significant overlap with any test examples from the training data, which raises a new issue: if the goal is to train a ``general-purpose'' model, what information could we safely exclude from training purely for evaluation purposes? 

\citet{Linzen_2020_How_Can_We_Accelerate_Progress_Towards_Human-like_Linguistic_Generalization} suggests switching to out-of-distribution testing: given that the training data is unlikely to faithfully represent a full range of linguistic phenomena, in-distribution evaluation likely overstates how well the model is doing. But to do that, we would still need to know what is ``in" the training distribution, and what we would be testing.

\vspace{1em}

To sum up, there are (at least) 4 reasons to make deliberate decisions about what should be included in the training data, so as to create more robust, inclusive, and secure NLP models. What are the objections?

\subsection{Why Not to Change the Data?}
\label{sec:con}

Since this is a position paper arguing that data curation is unavoidable, the arguments against it are presented together with the defense. Most of them are applicable to both pre-training and task data (except for \cref{con:large-enough}, which focuses on pre-training).

\subsubsection{Studying the world ``as it is".} \label{con:as-is} 
In response to \citeauthor{BenderGebruEtAl_2021_On_Dangers_of_Stochastic_Parrots_Can_Language_Models_Be_Too_Big}, \citet{Goldberg_2021_Criticism_of_Stochastic_Parrots} argued that there are valid use cases in which ``a model of language use should reflect how the language is actually being used", rather than how we believe it should be used.

\paragraph{Defense.} This is a completely valid argument, and what follows is elaboration rather than refutation. In linguistic or social science research, it is uncontroversial that if the corpus is a representative sample of the target phenomena, it should not be manipulated. If the goal is to model the worldview of Reddit users, the corpus used for training GPT-2 (comprising articles shared on Reddit) is a representative sample. Likewise, if the goal is to study social biases, we should not eliminate e.g. racist comments. The problem raised by \citet{BenderGebruEtAl_2021_On_Dangers_of_Stochastic_Parrots_Can_Language_Models_Be_Too_Big} is only that resources should be used for what they are: the Reddit users are not a representative sample of the general population, and so GPT-2 is not a ``general-purpose'' language model.

This argument concerns the qualitative studies of the ``world as it is''. Most NLP research, however, aims to produce systems that would perform some task. In that case the ``natural'' distribution may not even be what we want: e.g. if the goal is a question answering system, then the ``natural" distribution of questions asked in daily life (with most questions about time and weather) will not be helpful. The developers may also prefer for their systems to be e.g. less racist/sexist than their input data.

Note that to study the world ``as it is" we still have to do a lot more data work than we are currently doing (so as to be able to tell whether a given corpus actually represents the target phenomenon).

\subsubsection{Our sample is large enough.}
\label{con:large-enough}

An anonymous reviewer of this paper contributed the following argument: ``the size of the data is so large that, in fact, our training sets are not a sample at all, they are the entire data universe''.

\paragraph{Defense.} This argument would stand if the ``data universe" that we use for training NLP systems were the same as ``the totality of human speech/writing". It is not, and will hopefully never be, because collecting \textit{all} speech is problematic for ethical, legal, and practical reasons. Anything less than that is a sample. Given the existing social structures, no matter how big that sample is, it is not representative due to (at least) unequal access to technology, unequal possibility to defend one's privacy and copyright, and limited access to the huge volumes of speech produced in the ``walled garden" platforms like Facebook. The use of uncontrolled samples (like the Common-Crawl-based corpora) would have to be justified by arguing either that the above types of bias can be safely ignored, or that the benefits outweigh the risks.

\subsubsection{Might not be the best approach.} \label{con-algo}

Do we really have to do hard data work, or could there be an algorithmic solution? For the problem of rare phenomena (\cref{pro:long-tail}), there is ongoing work on inductive biases that could help the models learn them \cite{McCoyGrantEtAl_2020_Universal_linguistic_inductive_biases_via_meta-learning}. For social issues (\cref{pro:soc}) \citet{Goldberg_2021_Criticism_of_Stochastic_Parrots} and \citet{Buckman_2021_Fair_ML_Tools_Require_Problematic_ML_Models} similarly suggest that rather than trying to filter out problematic samples (hate speech, racial slurs etc.) we could use them to build a representation of the undesirable phenomena, and to try to actively identify and filter them out in generation. \citet{SchickUdupaEtAl_2021_Self-Diagnosis_and_Self-Debiasing_Proposal_for_Reducing_Corpus-Based_Bias_in_NLP} propose a method for a generative language model to reduce biases in its output, using self-diagnosis with its own internal knowledge. 

\paragraph{Defense.} It is entirely possible that algorithmic alternatives could work better than solutions based on data curation. Which one will be more successful is an empirical question. As of now, it seems that they are complementary rather than mutually exclusive: for example, some specific biases could be handled algorithmically, but data curation could be used to balance the corpus in some other way(s). 


Note that \textit{the algorithmic solutions would still require much of the same data work for evaluation purposes}: to find out whether a system is effective at filtering out something undesirable or processing some rare pattern, these phenomena have to be identified, a test set has to be constructed, we would need to make sure that it does not overlap with the training data, and ideally -- to what degree the various aspects of these phenomena are supported by training evidence. This is a big part of work that would go into designing a training dataset.

\subsubsection{Not what we set out to do!} The history of AI could be viewed as a trajectory towards decreased amount of implicitly injected knowledge. The early AI systems were fully driven by carefully constructed rules and ontoloties. They were replaced by the statistical approaches, relying on heavy feature engineering. The great promise of DL was to stop trying to define everything, and let the machine to identify and leverage patterns from huge datasets: ``we should stop acting as if our goal is to author extremely elegant theories, and instead embrace complexity and make use of the best ally we have: the unreasonable effectiveness of data" \cite{HalevyNorvigEtAl_2009_Unreasonable_Effectiveness_of_Data}. And it seems to work: pre-training larger models with more data keeps producing state-of-the-art results \cite{SunShrivastavaEtAl_2017_Revisiting_Unreasonable_Effectiveness_of_Data_in_Deep_Learning_Era,BrownMannEtAl_2020_Language_Models_are_Few-Shot_Learners,FedusZophEtAl_2021_Switch_Transformers_Scaling_to_Trillion_Parameter_Models_with_Simple_and_Efficient_Sparsity}.

Calls for careful construction of datasets are going in the face of that dream. We would arguably be even worse off than when we started: at least in the early AI days we only needed to define the phenomenon to be modeled, and now we also have to find hundreds of examples for that phenomenon.

\paragraph{Defense.} Disappointing as it is, we have to admit that although deep-learning-based systems are much better than their predecessors, they are still brittle and do not work well outside the range of cases well represented in the training data (and even there they may work for the wrong reasons). What is more, we are fundamentally no closer to the elusive idea of ``understanding" language or its meaningful production \cite{BenderKoller_2020_Climbing_towards_NLU_On_Meaning_Form_and_Understanding_in_Age_of_Data}. It is true that we were able to ``solve'' chess and Go without expert knowledge \cite{Sutton_2019_Bitter_Lesson}, but these are closed-world games with a known set of rules describing that world. Attempting to do so in the areas that feed from the real social world and impact that world (NLP, facial recognition, algorithmic decision-making on loans etc.) could amplify undesirable patterns present in the big data.

As stated in \cref{con-algo}, it is possible that there is an algorithmic approach that will work equally well or better. Which one will win is an empirical question. As of now, it is fair to say that data curation is at least an alternative to be considered.

This is \textit{not} to say that the current technology cannot yield useful solutions. The achievements are undeniable: the advances in machine translation, question answering, and dialogue already power better customer service, educate and inform, enable communication and information flow for people who could not afford professional translation. There is certainly room for useful research to further improve the current solutions, define new tasks and transfer to new domains and languages, even if no fundamental breakthroughs come any time soon. The question is only whether we want to be able to tell in what circumstances our models can be used safely \cite{MitchellWuEtAl_2019_Model_Cards_for_Model_Reporting}. If so, that would require more thinking about data.

\subsubsection{Perfection is not possible.} As mentioned in \cref{pro:long-tail}, the distribution of language phenomena tends to be Zipfian \cite{Zipf1945}, which means that most phenomena are rare and difficult to learn. A perfect dataset would provides a strong signal for each phenomenon that should be learned. That's not how language works, so we may never be able to create something like that. Balanced datasets are an improvement, but not a solution \cite{WangZhaoEtAl_2019_Balanced_Datasets_Are_Not_Enough_Estimating_and_Mitigating_Gender_Bias_in_Deep_Image_Representations,RogersKovalevaEtAl_2020_Getting_Closer_to_AI_Complete_Question_Answering_Set_of_Prerequisite_Real_Tasks}. 

\paragraph{Defense.} The impossibility of perfection does not entail the impossibility of improvement. For example, a sentiment analysis system that performs as well as the current systems \textit{while} handling negation and coreference correctly, and not pre-judging football fans as violent, is a doable next goal. 

\subsubsection{No single correct answer.} \label{con-wise} Curation means making conscious choices about what to include and what to exclude. These are essentially choices about \textit{designing a world}. What linguistic patterns, what concepts, what demographic attributes, what values should that world encode? This is a daunting question, requiring a lot of interdisciplinary expertise and impossible to casually address within a small NLP application project. Neither social sciences nor linguistics offer a ready set of answers, only things to consider in various contexts. The discriminated sub-groups, their values, and underlying social constructs may also differ across communities: e.g. both in India and US there is discrimination based on skin tone, but in the US context it stands for race, and in India it is a proxy for ethnicity, caste and class \cite{SambasivanArnesenEtAl_2021_Re-imagining_Algorithmic_Fairness_in_India_and_Beyond}.

\paragraph{Defense.} This is an entirely valid point, but it is an objection not to data curation \textit{per se}, but to ``data curation in a way that would inflict one set of values and linguistic choices on everyone". That is indeed to be avoided at all costs, and there is a real danger of that happening when NLP systems are commercially deployed and widely used, but the data choices behind them are not explicit.

The position advocated in this paper, as well as by \citet{BenderGebruEtAl_2021_On_Dangers_of_Stochastic_Parrots_Can_Language_Models_Be_Too_Big}, is only that whatever categories and demographics went into the data design, they have to be documented \cite{BenderFriedman_2018_Data_Statements_for_Natural_Language_Processing_Toward_Mitigating_System_Bias_and_Enabling_Better_Science,GebruMorgensternEtAl_2020_Datasheets_for_Datasets} and made explicit, so that the users could be informed about what is happening \cite{MitchellWuEtAl_2019_Model_Cards_for_Model_Reporting}. Some studies will just use convenience samples, and some will intentionally try to create a representation of a world without racial prejudice or rich with island effects. There are valid use cases for both, as long as it is clear who/what is being represented and for what purposes. The tide seems to be turning in this direction: since this work was submitted for review, at least two papers came out documenting popular resources for pre-training language models \cite{DodgeSapEtAl_2021_Documenting_English_Colossal_Clean_Crawled_Corpus,BandyVincent_2021_Addressing_Documentation_Debt_in_Machine_Learning_Research_Retrospective_Datasheet_for_BookCorpus}. The popular HuggingFace NLP dataset library\footnote{\url{https://huggingface.co/docs/datasets/}} is also working towards data cards for its resources.

Documenting the choices made in the dataset design is prerequisite to model cards \cite{MitchellWuEtAl_2019_Model_Cards_for_Model_Reporting}, which could facilitate a healthy interaction between the communities served by the system and the developers of that system. It is entirely possible for that interaction happen in a democratic process: the policies could developed, announced and updated based on the evolving user preferences. Robustness in handling linguistic and social peculiarities of a given community should be a selling point for a product striving to win that community over: something to compete for and showcase, rather than avoid mentioning.

When argument \cref{con-wise} is made, sometimes it seems to rest on the idea that the distributions in our resources objectively reflect the world. On that view, the calls to data curation would seem opinionated and unnecessary, if not outright dystopian. But the idea that it is possible to work on ``NLP in the vacuum", unmarked by linguistic and social categories, is an illusion. A decision to use a convenience sample is also a choice, an act of curation. Using any data to derive research conclusions or in commercial applications is only safe if we know what/who it represents.

\section{Why Curation Is Inevitable}
\label{sec:world}

In cognitive and sociolinguistics, one of the methods of studying the linguistic and conceptual repertoire of a certain individual or a demographic is through collecting a representative corpus of their speech (synchronic or diachronic). That corpus inevitably reflects a particular world view\footnote{This is a key concept in the works of Neo-Humboldtian scholars: ``world image'' (Weltbild) of \citeauthor{Weisgerber_1953_Vom_Weltbild_der_deutschen_Sprache}, ``naive picture of the world'' (naivnaja kartina mira) of \citet{Apresyan_1995_Izbrannyje_trudy}, and many others.}. The differences in these world views are expressed as variation in what kinds of linguistic structures people are likely to use, what they are likely to talk about, what are their presuppositions and social context and stereotypes, to what extent any of that is verbally expressed, etc. Some of that variation is idiosyncratic, some attributable to social groups, but even a cursory look at all the variation strongly suggests that there is no ``language in general".

It \textit{is} still possible to talk about language at a certain level of abstraction (e.g. ``British English" vs the myriad of UK dialects), but only with a good sample representing all the necessary subsets. For example, it would be wrong to construct a ``British English" resource based only on London samples, because they do not represent the rest of the country (either linguistically or socioeconomically).

A major achievement of corpus linguists are the ``national corpora" such as BNC \cite{Leech_1992_100_million_words_of_English_British_National_Corpus_BNC}, painstakingly created to represent a diverse sample of written and spoken genres in a certain geographical region in a certain timeframe, so as to enable studies of that specific variety of language. Creating such corpora involves careful sampling, detailed documentation of the domains and speakers that were represented, and much negotiation with publishers for copyright exceptions. 

A typical corpus for training language models, or really any NLP dataset, is likewise a sample of speech of a certain group of people, who have their linguistic preferences and sets of values. Consequently, that sample, whether it is coherent or not, and whether it was collected with any specific intentions, represents a certain ``picture of the world". Moreover, the purpose of using this data for training is to create a system that would encode that view of the world and make predictions consistent with it. But a typical NLP dataset\footnote{Corpora generated on crowd worker platforms such as Amazon Mechanical Turk typically impose geographic restrictions, such as ``location in US or Canada", but there is no guarantee that the recruited workers are even native speakers.} currently has few specifications of the demographics, dialects, or the range of domains and linguistic phenomena it covers. Unfortunately, it does not mean that the result is some abstract ``standard" or ``neutral" language. It is some kind of interpolation from the mixture of signals in the data that we have very little idea about. 

Why does it matter? The linguistic and conceptual repertoire of humans is dynamic. Our vocabulary, grammar, style, cultural competence change as we go on with our lives, encounter new concepts, forget some things and reinforce others. A key part of that change is the linguistic signals we encounter in communication: on the nativist account children have innate constraints that guide\footnote{The ``radical" nativist position would be that knowledge of language is entirely innate and is not affected by what the children observe, but on that position we would have to claim the innate knowledge of the word ``carburetor" \cite{Knight_2018_According_to_Chomsky_words_such_as_book_and_carburetor_are_genetically_determined}.} their learning from the data they encounter \cite{Chomsky_2014_Aspects_of_Theory_of_Syntax,HornsteinLightfoot_1985_Explanation_in_Linguistics_Logical_Problem_of_Language_Acquisition}, and on the usage-based accounts \cite{Bybee_2006_From_Usage_to_Grammar_The_Minds_Response_to_Repetition,LievenTomasello_2008_Childrens_first_language_acquistion_from_usage-based_perspective} that process is entirely data-driven. Humans can learn the meaning of words from a single exposure \cite{CareyBartlett_1978_Acquiring_Single_New_Word,BorovskyKutasEtAl_2010_Learning_to_use_words_Event-related_potentials_index_single-shot_contextual_word_learning}, but there is also robust evidence of frequency effects in language acquisition \cite{AmbridgeKiddEtAl_2015_ubiquity_of_frequency_effects_in_first_language_acquisition,DiesselHilpert_Frequency_effects_in_grammar}. It is not by accident that the frequency of the vocabulary to be learned is a key variable in language pedagogy \cite{ZaharCobbEtAl_2001_Acquiring_Vocabulary_through_Reading_Effects_of_Frequency_and_Contextual_Richness}. 

In short, humans, like DL models, learn from the patterns in the speech that they encounter. And those patterns do not have to come from human speakers anymore: much speech that we will encounter in the future is likely to be synthetic. According to \citet{Pilipiszyn_2021_GPT-3_Powers_Next_Generation_of_Apps}, GPT-3 is already generating 4.5B words per day in applications such as question answering, summarization, interactive games, and customer support. 

This cannot but have impact back on the human speakers\footnote{Synthetic speech will also clearly have impact on the future models if it seeps into the training data. There is research on watermarking generated text \cite{VenugopalUszkoreitEtAl_2011_Watermarking_Outputs_of_Structured_Prediction_with_application_in_Statistical_Machine_Translation,AbdelnabiFritz_2021_Adversarial_Watermarking_Transformer_Towards_Tracing_Text_Provenance_with_Data_Hiding}, but it is not clear what, if anything, the currently deployed systems are doing in this regard. There is at least one documented case of GPT-3 used to post on Reddit as if it were a human user \cite{Philip_2020_GPT-3_Bot_Posed_as_Human_on_AskReddit_for_Week}.} in the following ways: 

\begin{itemize*}
    \item An NLP system generating text contributes to a human learner's input in the same way as human writers, and probably also speakers (but potentially on a much larger scale). 
    \item An NLP system that processes human input to answer questions, translate, perform assisting actions etc. has both direct impact (as a language model above), and an indirect impact: as these systems become more widespread, the kind of language that they can and cannot successfully interpret will be respectively reinforced or made less prominent.
    \item An NLP system that makes decisions in processing applications, grading student work, curating news feeds, summarizing papers and emails, recommending content has the potential of making long-lasting impact on the lives of its users, and the kinds of language that it can process successfully clearly play a role. 
\end{itemize*}

The point to take from all of this is that any mismatch of linguistic and social feature distributions between NLP systems and their users \textit{will} have some impact on the world, and for the commercial, widely used NLP systems that impact may be significant. So the debate is not about whether we should change the world by making choices about the data: this is happening either way, because even our convenience samples still reflect numerous implicit choices. The debate is only about how much thinking we want to invest into changing our world.

This thought is somewhat scary (in what way will children growing up with Alexa be different?), but also exciting: the educational opportunities alone could be breathtaking, reaching far beyond the students who are already in a good position to do well in school. We could also create something simplistic, uninspiring, mindlessly entertaining, and/or not-inclusive. That choice is ours.

\section{What does it mean to ``do NLP"?}
\label{sec:interdisciplinary}

To sum up the above discussion: there are no ``neutral", one-size-fits-all textual corpora. There is also no manual that would provide foolproof instructions for collecting a ``correct" corpus for any given context. And all of these complications are not even the main problem, right? After all, data only serves the task of creating a model, which is the real contribution of an NLP paper?

In theory, the field of NLP is interdisciplinary. In practice, it became something closer to ``one of the applied areas of machine learning" rather than ``computational linguistics". Furthermore, at least as far as graduate students are concerned, it is something performed as an academic exercise, and as such it does not \textit{really} have to concern itself with its possible effects on the world. 

The students can hardly be blamed: keeping up with the latest frameworks and architectures is already hard enough. Most DL practitioners have neither the training nor time to also do the data work at the level that the linguists and ethicists are calling for. The publication system does not provide the right incentives for that either: modeling NLP work is prestigious and welcomed at top conferences, while data work is ``janitorial", less well paid, ``under-valued and de-glamorised''\footnote{Of course, this perception is not universal, and there are (very few) ``unicorn" resources like SQuAD \cite{RajpurkarZhangEtAl_2016_SQuAD_100000+_Questions_for_Machine_Comprehension_of_Text} that highly influenced the field. But overall the power balance in the field is currently not in favor of resource work.} \cite{SambasivanKapaniaEtAl_2021_Everyone_wants_to_do_model_work_not_data_work_Data_Cascades_in_High-Stakes_AI}. 

It does not help that there seems to be a systematic miscommunication between the fields. When linguists or ethicists talk about the issues with the current solutions, the practitioners may take it as an accusation that they are not doing a good job, rather than as an invitation to improve things together. Likewise, when the practitioners propose new systems, the linguists and ethicists may be frustrated: not by the incremental improvements on leaderboards as such, but by lack of accompanying discussion of what the proposed methods are supposed to do better, and for whom. 

If anything is to change, we need to overcome this antagonism. Here are a few suggestions for how that could be achieved.

\section{Moving Forward}
\label{sec:solutions}

\paragraph{Step 1. Understand each other better.} 

The fact is, the AI ethics people are not really out to ``cancel" everybody. It is easy to see why they would be frustrated that the social justice issues have never been a priority, terrified at what ``move fast \& break things" has already done with the social world, and dubious that they just need to wait and change would come.

The linguists are not completely useless. Chances are, many problems that the DL engineers are having could be fixed if someone was just around to realize that the tokenizer didn't handle the suffixes well.

And the engineers are not inherently evil. They just need resources, training, collaborators, time, and better research incentives. Instead, they have to churn out papers in 2 months just to stay in the publication race, with no time to dive deeper into what their systems are actually doing.

\paragraph{Step 2. Improve the incentive structure.}

One way to change the incentive structure that led to the current situation is through conferences. There will be a lot more interest in data work if it becomes more publishable. As of now, the ``resources and evaluation'' track is something of a poor relative to the ``machine learning'' track, which in ACL 2020\footnote{\url{https://www.aclweb.org/adminwiki/images/9/90/ACL_Program_Co-Chairs_Report_July_2020.pdf}} attracted nearly 3 times more submissions. Most task-specific tracks (question answering, summarization, dialogue etc.) are supposed to receive both engineering and data submissions, but in that setting the interdisciplinary tension may lead to resource papers voted down simply for being resource papers \cite{RogersAugenstein_2020_What_Can_We_Do_to_Improve_Peer_Review_in_NLP}. \citet{Bawden_2019_One_paper_nine_reviews} cites an ACL 2019 reviewer who complained that ``the paper is mostly a description of the corpus and its collection and contains little scientific contribution". 

We really need to take the type of contribution\footnote{As was done e.g. at COLING 2018: \url{http://coling2018.org/paper-types/}} into account in reviewer assignment, into review form design, and into reviewer training programs. We also need to make sure that the resource tracks are consistently offered\footnote{E.g. this track was recently absent at EMNLP 2020.}, with dedicated best paper awards to raise the prestige of this work in the community. Some conferences already started to provide reviewer mentoring, double down on ethics, consider what signal they send to companies and students by their best paper awards. We can all help by lobbying program chairs whenever we have a chance, offline and online.

A helpful factor is that the ever-increasing size of models is making the state-of-the-art leaderboard chase financially untenable for even well-resourced labs, and they are looking for other outlets. This is a chance for the NLP community to engage more deeply with the phenomena that we are modeling.

\paragraph{Step 3. Educate.}

The idea that ``NLP" means ``deep learning" may well arise if it is taught as a one-semester course focusing on the engineering. If the coursework is fully powered by existing resources, it creates the impression that data is not a part of the job. The result is that the students learn that it is entirely possible to just run off-the-shelf parsers without knowing anything about syntax, or do sentiment analysis without knowing anything about pragmatics. And if it is possible to not do more work, why would anyone bother?

We need to provide our students with the skills to stress-test their systems and critically examine their data, so as to be able to spot potential issues early on. For that, they will need the basic linguistic theory, the fundamentals of sociolinguistics and pragmatics. Likewise, some aspects of psychology (dual processing theories, memory and attention span, cognitive biases, ``nudging") are a pre-requisite for designing interfaces not only for annotation projects, but for any kind of interactive NLP systems. And some awareness of the social power structures would help in not propagating the harmful stereotypes. Some strategies for building NLP curricula have been discussed at the TeachingNLP workshop \cite{RadevBrew_2002_Proceedings_of_ACL-02_Workshop_on_Effective_Tools_and_Methodologies_for_Teaching_Natural_Language_Processing_and_Computational_Linguistics,BrewRadev_2005_Proceedings_of_Second_ACL_Workshop_on_Effective_Tools_and_Methodologies_for_Teaching_NLP_and_CL,PalmerBrewEtAl_2008_Proceedings_of_Third_Workshop_on_Issues_in_Teaching_Computational_Linguistics,DerzhanskiRadev_2013_Proceedings_of_Fourth_Workshop_on_Teaching_NLP_and_CL,JurgensKolhatkarEtAl_2021_Proceedings_of_Fifth_Workshop_on_Teaching_NLP}.

Most importantly, NLP courses need to combat the idea that all the knowledge about the human world is just irrelevant in the age of big data and DL. The ``garbage in, garbage out'' principle is still relevant. We may be able to sort the garbage and learn from it anyway, but only if we have at least some idea about what kind of garbage we have.

\paragraph{Step 4. Collaborate.}

Large companies and universities provide a significant competitive edge to their authors just in virtue of the in-house collaboration networks they could offer. But it is becoming increasingly easy for everyone to find external collaborations, especially in the world in pandemic lockdown. One opportunity is Twitter, used by estimated 40\% of EMNLP 2020 authors\footnote{Source: EMNLP 2020 organizers.}.

What would it mean to ``collaborate"? At the bare minimum, in an engineering project the linguists and social scientists could help to at least try to characterize the data that was used with something like data statements \cite{BenderFriedman_2018_Data_Statements_for_Natural_Language_Processing_Toward_Mitigating_System_Bias_and_Enabling_Better_Science,GebruMorgensternEtAl_2020_Datasheets_for_Datasets}. A more ambitious goal would be to involve them early on in the data selection, preparation, and iterative development. Ideally, there would be joint formulation of research goals, thinking together about what kind of world we are building.

Finding collaborators is much easier for established researchers, not only because they are a known quantity, but also because they are already aware of what could be done in an interdisciplinary project. They probably even already know the people who they could ask to join. But the students could use some help, especially those from the less well-connected institutions. They could benefit from establishing some kind of skill exchange network, where the students with engineering background could help out in data projects and students with linguistics/social science background could help out in engineering projects. This would probably the best way to ease the interdisciplinary tension, instill respect for each other's expertise, as well as the awareness that NLP is a huge problem that we do not even understand that well, and for which we need all the help we can get.

\paragraph{Step 5. Estimate.}
The goal of all the above data work is ultimately to enable informed decisions by the public, the CEOs, and the policy makers about what kind of world we would live in. One takeaway from the heated debate around \cite{BenderGebruEtAl_2021_On_Dangers_of_Stochastic_Parrots_Can_Language_Models_Be_Too_Big} is that if one side in an interdisciplinary debate focuses mostly on the potential benefits of something, and the other mostly on its harms, the stance is likely to become adversarial, and we do not give each other the benefit of the doubt\footnote{\url{https://twitter.com/nlpnoah/status/1354814467633111048}}. 

Nevertheless, the people on both sides of the debate are researchers, and they want to make informed decisions. That is only possible through cost-benefit analysis. It is clear that the first step has to be thorough documentation of the data \cite{BenderFriedman_2018_Data_Statements_for_Natural_Language_Processing_Toward_Mitigating_System_Bias_and_Enabling_Better_Science,GebruMorgensternEtAl_2020_Datasheets_for_Datasets}: this lets us compare the represented population and the population of the target users, and think through the possible harms. However, it is not clear how to weigh the harms against the benefits. 

At the very least, to make informed decisions we would probably need to know the following\footnote{Many of these points are made in the NAACL ethics FAQ \url{https://2021.aclweb.org/ethics/Ethics-FAQ/}}:

\begin{itemize*}
    \item Which population will get exposed to the proposed tech?
    \item What are the direct and indirect benefits on the user population?
    \item What are the direct and indirect harms on the population in general (not limited to the users of the proposed tech), in particular the marginalized groups?
    \item If certain harms are inflicted on the user population, would they have the political/legal recourse to be compensated?
    \item How compute-efficient the implementation would be, how would the energy be sourced, and would that affect any other populations?
    \item How widely would it be eventually adopted, and how that changes the likelihood of benefits and harms to different user groups?
    \item What is the potential for further innovation that would significantly change the appeal, deployability or risks of the proposed solution?
    \item What are the risks of human error and deliberate misuse if the tech is stolen/replicated by terrorists, authoritarian governments, propaganda organizations and other bad actors?
\end{itemize*}

Unfortunately, the world is volatile and business plans change all the time. There is so much uncertainty for each of these points that it is not clear how to even start. Yet we have to try to come up with a process for working these things out, and eventually develop templates and calculators that developers could use to make estimates for best-, worst- and realistic scenarios. 

This is an area in which NLP is desperately in need of collaboration with economics, governance and law. In that, again, NLP conferences could take the lead. There could be regular tracks that would incentivize joint publications with experts from these fields. The search for solutions is already going on, but this way NLP community would participate in it rather than just meet with regulation post-factum. To be able to provide meaningful peer review for such work, we would need a mechanism of recruiting external reviewers with the required expertise on as-need basis.

\section{Conclusion}

Our data is already changing the world, and will keep doing so whether we are being intentional about it or not. We might as well at least try: we do want more robust and linguistically capable models, and we do want models that do not leak sensitive data or propagate harmful stereotypes. 

Whether those goals would be ultimately achieved by curating large corpora or by more algorithmic solutions, in both cases we need to do a lot more data work. The current dynamic suggests that this won't happen, unless we overcome the interdisciplinary tensions and turn our conferences into truly shared spaces. 

\section{Acknowledgements}

Many thanks to Emily M.~Bender, Yoav Goldberg, Ryan Cotterell, and the anonymous reviewers for their thoughtful comments on this paper.

\bibliographystyle{acl_natbib}
\bibliography{acl2021}

\end{document}